\documentclass{bmvc2k}

\title{Open-vocabulary Semantic Segmentation with Frozen Vision-Language Models}

\addauthor{Chaofan Ma$^*$}{chaofanma@sjtu.edu.cn}{1}
\addauthor{Yuhuan Yang$^*$}{yangyuhuan@sjtu.edu.cn}{1}
\addauthor{Yanfeng Wang$^\dagger$}{wangyanfeng@sjtu.edu.cn}{1,2}
\addauthor{Ya Zhang}{ya_zhang@sjtu.edu.cn}{1,2}
\addauthor{Weidi Xie}{weidi@sjtu.edu.cn}{1,2}

\addinstitution{
Coop. Medianet Innovation Center,\\
Shanghai Jiao Tong University, China
}
\addinstitution{
Shanghai AI Laboratory
}
\runninghead{MA, ET.AL.}{Open-vocabulary Semantic Segmentation with Frozen VLMs}

\def\ie{\emph{i.e}\bmvaOneDot}
\def\eg{\emph{e.g}\bmvaOneDot}

\definecolor{vis_red}{rgb}{0.918,0.263,0.208}
\definecolor{vis_green}{rgb}{0.204,0.659,0.325}
\definecolor{vis_blue}{rgb}{0.259,0.522,0.957}
\definecolor{vis_yellow}{rgb}{0.984, 0.737, 0.016}
\definecolor{bmvc_blue}{RGB}{0,0,102}

\newcommand{\figref}[1]{Figure~\ref{#1}}
\newcommand{\tabref}[1]{Table~\ref{#1}}
\newcommand{\secref}[1]{Section~\ref{#1}}

\usepackage{bm}  
\usepackage{float}
\usepackage{times}
\usepackage{bbding}
\usepackage{xspace}  %
\usepackage{dsfont}
\usepackage{epsfig}
\usepackage{amsmath}
\usepackage{amssymb}
\usepackage{comment}
\usepackage{graphicx}
\usepackage{booktabs}
\usepackage{multirow}
\usepackage{colortbl}
\usepackage{enumitem}
\usepackage{placeins}
\usepackage{footnote}
\usepackage{multicol}
\usepackage{graphbox}
\usepackage{placeins}
\usepackage{mathrsfs} 
\usepackage{enumitem}
\usepackage{makecell}
\usepackage{threeparttable}

\usepackage{caption}
\newfloat{figtab}{htb}{fgtb}
\makeatletter
  \newcommand\figcaption{\def\@captype{figure}\caption}
  \newcommand\tabcaption{\def\@captype{table}\caption}
\makeatother

\usepackage{algorithmic}
\usepackage{algorithm}
\definecolor{commentcolor}{RGB}{110,154,155}   

\usepackage{hyperref}

\begin{document}

\maketitle
\def\thefootnote{}\footnotetext{$*$ Equal contribution. $\dagger$ Corresponding author.}
\def\thefootnote{\arabic{footnote}}

\begin{abstract}
When trained at a sufficient scale, 
self-supervised learning has exhibited a notable ability to solve a wide range of visual or language understanding tasks.
In this paper, we investigate simple, yet effective approaches for adapting the  pre-trained foundation models to the downstream task of interest, 
namely, open-vocabulary semantic segmentation.
To this end, we make the following contributions:
(i) we introduce \textbf{Fusioner}, with a lightweight, transformer-based fusion module, that pairs the {\em frozen} visual representation with language concept through a handful of image segmentation data. As a consequence, 
the model gains the capability of zero-shot transfer to segment novel categories;
(ii) without loss of generality, 
we experiment on a broad range of self-supervised models that have been pre-trained with different schemes, {\em e.g.}~visual-only models~(MoCo v3, DINO), language-only models~(BERT), visual-language model~(CLIP), and show that, the proposed fusion approach is effective to any pair of visual and language models, even those pre-trained on a corpus of uni-modal data;
(iii) we conduct thorough ablation studies to analyze the critical components in our proposed \textbf{Fusioner}, while evaluating on standard benchmarks, {\em e.g.}~PASCAL-5$^i$ and COCO-20$^i$, 
it surpasses existing state-of-the-art models by a large margin, despite only being trained on frozen visual and language features;
(iv) to measure the model's robustness on learning visual-language correspondence, we further evaluate on a synthetic dataset, named Mosaic-4, where images are constructed by mosaicking the samples from FSS-1000. \textbf{Fusioner} demonstrates superior performance over previous models. 
Project page: \href{https://yyh-rain-song.github.io/Fusioner_webpage/}{https://yyh-rain-song.github.io/Fusioner\_webpage/}.
\end{abstract}

\section{Introduction}

In the recent literature, self-supervised representation learning has made remarkable progress. 
For example, MoCo~\cite{mocov3} and DINO~\cite{dino} have shown the possibility of learning strong visual representation without using any manual annotation. 
Noticeably, despite being trained purely by self-supervised learning on images, 
these image models have shown to implicitly capture the concept of objectness, 
{\em i.e.}~free semantic segmentation to some extent~\cite{Melaskyriazi22, Shin22, Wang22tokencut}.
Another line of work proposes to learn joint representation for image and text on very large-scale image-text pairs collected from Internet. 
For example, by training with simple noise contrastive learning,
CLIP~\cite{clip} and ALIGN~\cite{align} have demonstrated impressive ``zero-shot'' transferability and generalizability in various image classification tasks. 
With the growing computation, 
it is thus foreseeable that more powerful models will be trained, with even larger scale datasets,  
further pushing the performance on image classification.

In contrast, semantic segmentation considers a more challenging task, 
that aims to assign one semantic category to each pixel in an image.
By training deep neural networks on large-scale segmentation datasets, 
recent models have shown great success, for example, 
FCN~\cite{FCN}, U-Net~\cite{UNet}, DeepLab~\cite{DeepLab}, DPT~\cite{DPT}.
However, 
the conventional strategy that trains parametric classifiers for individual category, has posed fundamental limitations,  
as it only allows the model to make predictions on a close-set of predetermined categories at inference time, \ie, the model can only segment objects of the training set classes,
and lacks the ability to handle samples from novel~(unseen) classes.

In this paper, 
we explore efficient ways to bridge the powerful pre-trained vision-only, language-only 
or visual-language models, which, as a result, effectively tackles the problem of \textbf{open-vocabulary semantic segmentation}, \ie, segmenting objects from any categories by their textual names or description.
In specific, we first encode the input image and category names with {\em frozen}, self-supervised visual and language models. 
The computed features are then concatenated and passed to a transformer-based fusion module, that enables the features to be iteratively updated, in condition of the other modality through self-attention. 
As to obtain the segmentation mask, 
we further measure the cosine similarity between each pixel and the textual features of each category,
\ie, computing dot product between the L2-normalised visual and language embeddings.

To summarize, we make the following contributions: 
{\em First}, to tackle the problem of open-vocabulary semantic segmentation, we introduce \textbf{Fusioner} with a simple, lightweight transformer-based fusion module, which enables to explicitly bridge powerful, pre-trained visual, language, or visual-language models;
{\em Second}, we demonstrate the idea's effectiveness by experimenting on a wide spectrum of self-supervised models, that are pre-trained with completely different schemes, for example, MoCo v3, DINO, CLIP, BERT, 
and show that the proposed fusion approach is effective to any pair of visual and language models, even those pre-trained on a corpus of uni-modal data;
{\em Third}, we conduct thorough ablation studies to validate the critical components. Despite all visual and language models are kept frozen, the proposed simple fusion module is able to outperform the existing state-of-the-art approaches in ``zero-shot'' settings significantly,
and is competitive across numerous few-shot benchmarks, \eg, ~PASCAL, COCO, FSS-1000;
{\em Fourth}, to measure the model's robustness for learning visual-language correspondence, 
we introduce a new dataset, named Mosaic-4, that can be used in open-vocabulary semantic segmentation to detect whether the models tend to segment saliency that ignore the textual input. \textbf{Fusioner} shows superior performance over previous models.  

\section{Related Work}

{\noindent \bf Pre-trained Vision and Language Models.}
Self-supervised representation learning has recently made substantial progress in both computer vision and natural language processing.
In specific, some pre-training methods adopt contrastive learning (SimCLR \cite{simclr}, MoCo \cite{mocov3}, SwAV \cite{swav}), metric-learning (BYOL \cite{byol}, SimSiam \cite{simsiam}) and self-distillation (DINO \cite{dino}) can all be seen as powerful feature extractors on downstream tasks. On the other hand, language model pre-training may use a masked language modeling loss (BERT \cite{bert}, T5 \cite{t5}) or next-token prediction loss (GPT \cite{gpt1, gpt2, gpt3}). 
Not surprisingly, large-scale visual-language models have also attracted growing attention~\cite{lu2019vilbert, su2019vl, chen2020uniter, li2020unicoder, li2020oscar, lei2021less},
a milestone work is Contrastive Language-Image Pre-training (CLIP)~\cite{clip},
that trains on the large-scale image-text pairs with simple noise contrastive learning, 
and has shown strong capability of aligning two modalities in embedding spaces.
Inspired by this work, a series of studies have been proposed to transfer the knowledge of the pre-trained CLIP and extend to various downstream tasks. For example, object detection \cite{vild,feng2022promptdet}, image captioning \cite{hessel2021clipscore}, referring image segmentation \cite{wang2022cris}, text-driven image manipulation \cite{patashnik2021styleclip}, and supervised dense prediction \cite{rao2022denseclip}, etc.
Unlike these works that {\em fine-tune} CLIP for different downstream tasks,
we explore an alternative approach, and show that, 
simply pairing {\em any} frozen self-supervised visual and language models with lightweight fusion module, can be a surprisingly strong baseline for open-vocabulary semantic segmentation.\\[-8pt]

{\noindent \bf Bridging Pre-trained Models.}
With the rapid development of foundation models~\cite{bommasani2021opportunities, yuan2021florence, chowdhery2022palm}, 
many works have studied effective ways to adapt different downstream tasks by composing different pre-trained models. Frozen \cite{frozen} proposed to align the pre-trained, 
frozen language model with vision encoder by learning continuous prompts with only a few examples.
Flamingo \cite{flamingo} proposed an architecture that uses large pre-trained vision-only and language-only models to learn a wide range of visual-language task with only limited samples.
Socratic Models \cite{socratic}, which uses pre-trained language models, vision-language models, and audio-language models to complete downstream multi-modal tasks, such as image captioning and video-to-text retrieval, without the need of training.\\[-8pt]

{\noindent \bf Semantic Segmentation of Novel Categories.}
To enable a network for segmenting novel categories is still an open and active research question, as most of the existing semantic segmentation methods are limited to a closed set, \ie, the category of test set is the same as the training set.
Zero-shot semantic segmentation often take advantage of category-level semantic word embedding to segment novel categories without additional samples. For example, ZS3Net \cite{zs3net}, CSRL \cite{csrl}, CaGNet \cite{gu2020context}, and CaGNet-v2 \cite{gu2022pixel} are generative methods combining a deep visual segmentation model with an approach to synthesize visual features for novel categories based on semantic word embeddings. 
SPNet \cite{spnet}, JoEm \cite{baek2021exploiting}, LSeg~\cite{lseg} are discriminative methods mapping each pixel and semantic word to a joint embedding space, and leveraging the joint embedding space to give the class probability.
Our approach falls into the latter line, however, 
we advocate a lightweight fusion module that only aligns the pre-trained, frozen visual and language features.

\section{Methods}

In this section, we start by formulating the the problem of open-vocabulary semantic segmentation, and then detail our proposed architecture, termed as \textbf{Fusioner}, 
that addresses the task by bridging the pre-trained vision and language models. \\[-0.8cm]

\paragraph{Problem Formulation.}
Following the same setting as defined in LSeg~\cite{lseg},
we are given a training set $\mathcal{D}_{\text{train}}=\{
(\mathcal{X}, \mathcal{Y}, \mathcal{C})|
\mathcal{X}\in\mathbb{R}^{H \times W \times 3},
\mathcal{Y}\in\mathbb{R}^{H \times W \times |\mathcal{C}|},
\mathcal{C} \subseteq \mathcal{S} \}$, 
where $\mathcal{X}$ denotes any input image;
$\mathcal{Y}$ refers to the segmentation masks for $\mathcal{X}$ with one-hot encoding;
$\mathcal{C}$ refers to a set of seen categories in $\mathcal{X}$; $\mathcal{S}$ is all training (seen) categories in this training set.
Our goal here is to train a segmentation model that can partition a test image into semantically meaningful regions of {\em unseen} categories:
\begin{align}
  \mathcal{Y}_j = \Phi_{\textsc{Fusioner}}\left(\mathcal{X}_j, \mathcal{W}_j \right) =\Phi_{\textsc{Dec}}(\Phi_{\textsc{Fuse}}(\Phi_{\textsc{Visual-Enc}}(\mathcal{X}_j), \Phi_{\textsc{Text-Enc}}(\mathcal{W}_j))),
\end{align}
where $\mathcal{W}_j=\{w_j^1, \dots, w_j^{|\mathcal{W}_j|}\}$ is the categories of interests in one image  $\mathcal{X}_j$ ($|\mathcal{W}_j|$ textual words, \eg, ``cat'', ``plant''), and is dynamic for different images. The corresponding $|\mathcal{W}_j|$ output binary segmentation maps denote as $\mathcal{Y}_j \in\mathbb{R}^{H \times W \times |\mathcal{W}_j|}$.
Note that, for the conventional close-set segmentation, $\mathcal{W}_j \subseteq \mathcal{S}$, 
while for the open-vocabulary problem, $\mathcal{W}_j \bigcap \mathcal{S} = \emptyset$,
\ie, we evaluate the performance on the novel (unseen) categories that do not appear in the training categories $\mathcal{S}$.

\begin{figure}[t]
\centering
\includegraphics[width=.95\textwidth]{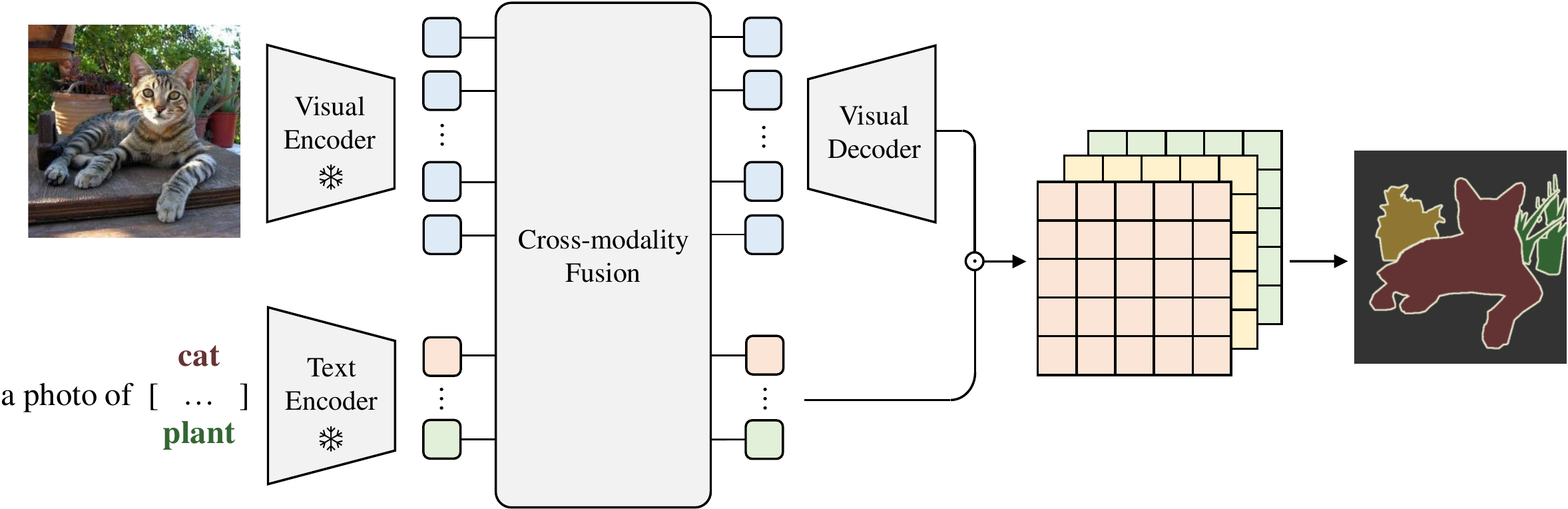}
\caption{Overview of our proposed \textbf{Fusioner}, which consists of a frozen visual and text encoder, a cross-modality fusion module and a visual decoder. The frozen encoders extract features for different modalities, and the fusion module bridges these embedding spaces. After upsampling the visual feature to its original resolution by the visual decoder, segmentation can be acquired by simple computing the similarity between the visual and language modalities.} 
\label{fig:framework}
\end{figure}

\subsection{Architecture}
\label{sec:arch}

The overall framework of our proposed \textbf{Fusioner} is illustrated in Figure~\ref{fig:framework}. It consists of three components: 
visual and language encoders for extracting features~(Section~\ref{sec:VL-Embedding}); 
a cross-modality fusion module to bridge the embedding spaces~(Section~\ref{sec:Fuse}); 
and an image decoder that upsamples the visual features to facilitate segmentation on original resolution as input images, and segmentation can be acquired by simple computing the cosine similarity between the visual and language~(Section~\ref{sec:decoder}).

\subsubsection{Visual and Language Representation}
\label{sec:VL-Embedding}

Here, we adopt pre-trained vision and language models as our encoders, 
and keep them frozen during training.
In specific, we mostly consider the transformer-based architectures, 
due to their good performance, and flexibility for encoding signals of different modalities. \\[-0.7cm]

\paragraph{Visual Feature Embeddings.} 
As the input to a vision transformer~\cite{vit}, 
the image $\mathcal{X}\in \mathbb{R}^{H \times W \times 3}$ is first split into a set of 2D non-overlapping patches, and projected into a sequence of vector embeddings. 
After adding positional embeddings~(inherited from the pre-trained vision transformer), 
these tokens are further processed by a series of transformer encoder layers,
each layer consists of multi-head self-attention (MHSA) and feed-forward network (FFN) together with layer normalization and residual connections:
\begin{align}
    \mathcal{F}_{\textsc{v}} =  \Phi_{\textsc{Visual-Enc}}(\mathcal{X}) \in \mathbb{R}^{(hw) \times d_{i}},
\end{align}
where $h={H}/{p}$ and $w={W}/{p}$, 
$p$ is the patch size, and $d_{i}$ is the visual embedding dimension. 
In later experiments, we adopt various popular transformer-based image encoders,
that were pre-trained with different self-supervised learning regime, 
for example, MoCo v3~\cite{mocov3}, DINO~\cite{dino}, and CLIP~\cite{clip}.\\[-0.7cm]

\paragraph{Textual Feature Embeddings.} 
The text embeddings are generated by encoding the semantic categories $\mathcal{W}$ through a text encoder:
\begin{align}
    \mathcal{F}_{\textsc{w}} =  \Phi_{\textsc{Text-Enc}}(\mathcal{W}) \in \mathbb{R}^{|\mathcal{W}| \times d_{w}},
\end{align}
where $|\mathcal{W}|$ refers to the input number of categories, and $d_w$ is the dimension of word embeddings. 
Prior to feeding the semantic category into text encoder, 
we use multiple prompt templates as decorations, \eg, ``a photo of \{category\} in the scene'', and average the output embeddings from text encoder. 
The complete prompt templates are listed in supplementary. 
We consider various self-supervised language models that were trained on a large corpus of documents or images as the text encoder, for example, BERT~\cite{bert}, or CLIP~\cite{clip}. 

\subsubsection{Cross-modality Fusion}
\label{sec:Fuse}

Given the visual features $\mathcal{F}_{\textsc{v}}$ and textual features $\mathcal{F}_{\textsc{w}}$, we firstly unify the channel dimensions for both visual and textual embeddings by using MLPs, {\em i.e.}~$\mathcal{F}_{\textsc{v}} \in \mathbb{R}^{(h w) \times d}$, $\mathcal{F}_{\textsc{w}} \in \mathbb{R}^{|\mathcal{W}| \times d}$,
and pass the resulting features through a cross-modality fusion module to adaptively capture the interactions between visual and language signals:
\begin{align}
    [\mathcal{F}_{\textsc{v}}', \text{ } \mathcal{F}_{\textsc{w}}'] = \Phi_{\textsc{Fuse}}([\mathcal{F}_{\textsc{v}}, \text{ } \mathcal{F}_{\textsc{w}}]),
\end{align}
where $[\cdot, \cdot]$ indicates feature concatenation of the visual and textual sequence.
$\Phi_{\textsc{Fuse}}$ is consisted of multiple transformer encoder layers, 
effectively capturing the long-range dependencies between the images and associated texts. 
The multi-modality visual feature $\mathcal{F}_{\textsc{v}}'$ and textual feature $\mathcal{F}_{\textsc{w}}^{\prime}$ have the same shape as $\mathcal{F}_{\textsc{v}}$ and $\mathcal{F}_{\textsc{w}}$, and both features are enriched and refined by iteratively attending the other modality. 

\subsubsection{Visual Decoder}
\label{sec:decoder}

\paragraph{Modality-maintained Upsampling.} 
Here, the visual features are progressively upsampled to the same resolution as the original image,
in detail, we first reshape the sequence of visual vectors into a spatial feature map, 
and upsample it by alternating convolutional and upsampling layers,
obtaining high resolution feature maps, 
\ie, $\mathcal{F}_{\textsc{v}}' \in \mathbb{R}^{{H} \times {W} \times d}$.\\[-0.8cm]

\paragraph{Calculating Segmentation Mask.}
The logits $\hat{y}$ is generated by computing the cosine similarity  between the upsampled feature map and textual feature, \ie, 
$\hat{y}=\mathcal{F}_{\textsc{v}}' \cdot  \mathcal{F}_{\textsc{w}}'^{\mathrm{T}} \in \mathbb{R}^{H \times W \times |\mathcal{W}|}$,
where $\mathcal{F}_{\textsc{v}}' \in \mathbb{R}^{{H} \times {W} \times d}$, $\mathcal{F}_{\textsc{w}}' \in \mathbb{R}^{|\mathcal{W}| \times d}$, 
denoting the L2-normalised visual and textual features,
and $|\mathcal{W}|$ is the number of categories (textual words). 
Seen as binary segmentation for each category, the final predictions can be obtained by simply applying sigmoid with a temperature $\tau$ and threshold classwise.

\subsubsection{Discussion} 
The closest work to ours is LSeg~\cite{lseg}, 
which also considers to tackle the problem of open-vocabulary semantic segmentation, 
by explicitly pairing high-capacity image and text encoder, however, 
there remains \textbf{three} critical differences, 
in LSeg, 
(i) the visual model (dense prediction transformers~\cite{DPT}) is pre-trained with supervised learning, while we use self-supervised models that can be easily scaled up;
(ii) the visual model is optimised end-to-end for segmentation on certain categories,
which may potentially lead to the catastrophic forgetting;
(iii) the visual and language representation are computed independently with dual encoders, 
and only fused at the last layer for computing semantic segmentation, 
thus referring to as a \textbf{late fusion}.
Such late fusion can potentially suffer from lexical ambiguities,
for example, same word (synonym) may refer completely different visual patterns,
while \textbf{early fusion} (ours) allows to update the features in condition to the other, 
potentially enabling to learn better visual-language correspondence.
In \secref{sec:exp}, we have conducted experiments to validate the superiority of early fusion.

\section{Experiment}
\label{sec:exp}

\subsection{Experimental Setups}
\label{Experimental Setups}

\paragraph{Datasets.} 
In accordance to prior work on open-vocabulary semantic segmentation~\cite{lseg},
we also evaluate our model on two benchmarks: PASCAL-5$^i$ and COCO-20$^i$. 
PASCAL-5$^i$ is the extension of PASCAL VOC 2012~\cite{everingham2010pascal} 
with extra annotations from SDS~\cite{hariharan2011semantic}, 
consisting of 20 semantic categories that are divided evenly into 4 folds containing 5 classes each, \ie, $\{5^i\}_{i=0}^3$. 
COCO-20$^i$ is built from MS COCO~\cite{lin2014microsoft} and contains 80 semantic categories that are also divided into 4 folds, \ie, $\{20^i\}_{i=0}^3$, with each fold having 20 categories. 
For each dataset, we conduct 4-fold cross-validation with same hyperparameter setup.

In addition, for robustness test, 
we introduce a new dataset with images constructed by mosaicking images from FSS-1000~\cite{fss1000}, termed as Mosaic-4.
FSS-1000 contains pixel-wise annotation of 1000 classes with 10 object-centric images each, in which $240$ classes ($2400$ images) are reserved for test.
Mosaic-4 reorganizes the test split of FSS-1000 by randomly sampling 4 images of different categories without replacement, and mosaicking them into one, creating a test list of $600$ compound images with explicit distractors. An example can be seen in \figref{fig:fss4} (a) and (b), where each color represents an individual category. \\[-0.8cm]

\paragraph{Implementation Details.} 
We experiment with three different pre-trained vision/language models:  
the vision-only model (MoCo v3, DINO), the language-only model (BERT), 
and the vision-language model (CLIP). 
\textbf{Note that}, all of them were kept frozen during training. 
The cross-modality fusion module contains 12 layers with 8 heads, and the visual decoder consists of $k$ layers of convolution followed by a $2\times$ upsampling, where $k=4$ for ViT backbone and $k=5$ for ResNet. 
We adopt AdamW optimizer, and the learning rate is ramped up during the first 10 epochs to 0.001 linearly. After this warmup, we decay the learning rate with a cosine schedule. The temperature factor $\tau=0.07$, and cross entropy is used for training. \\[-0.8cm]

\paragraph{Evaluation Metrics.} 
We adopt class mean intersection-over-union (mIoU) as our main evaluation metric, 
The mIoU averages IoU over all classes in a fold: $\mathrm{mIoU}=\frac{1}{C} \sum_{c=1}^{C} \operatorname{IoU}_{c}$, where $C$ is the number of classes in the target fold, 
and $\operatorname{IoU}_{c}$ is the intersection over union of class $c$.
We also consider FB-IoU in some experiments, however, FB-IoU only cares about the performance on target and non-target regions instead of differentiating categories, where only the foreground and background are considered as two categories ($C=2$).

\subsection{Ability to Bridge Different Pre-trained Backbones}
\label{Backbones Pairing Ability}

\begin{table}[!htb]
    \centering
    \footnotesize{}
    \setlength\tabcolsep{5pt}
    \resizebox{.95\textwidth}{!}{%
    \begin{tabular}{c|cc|cccc|c|cccc|c}
    \toprule
    \multirow{2}{*}{Model} &
    \multirow{2}{*}{\begin{tabular}{c}Visual\\Encoder\end{tabular}}&\multirow{2}{*}{\begin{tabular}{c}Text\\Encoder\end{tabular}}&\multicolumn{5}{c|}{Early Fusion}&\multicolumn{5}{c}{Late Fusion}\\
    \cmidrule(l){4-13}
    &&& $5^0$ & $5^1$ & $5^2$ & $5^3$ & mIoU & $5^0$ & $5^1$ & $5^2$ & $5^3$ & mIoU \\
    
    \midrule
    \multirow{3}{*}{A} & 
    \multirow{3}{*}{CLIP-B} 
     & CLIP-B & 50.2&62.4&\textbf{51.5}&44.4&52.1&46.3 &55.2 &36.4 &36.7 &43.6\\
     & & BERT-B & 46.6&61.3&44.6&43.7&49.1&47.5 &58.5 &38.5 &39.1 &45.9\\
     & & BERT-L &48.5&59.3&47.5&43.8&49.7&47.0 &56.6 &41.4 &38.8 &46.0\\
     
     \midrule
     \multirow{3}{*}{B} & \multirow{3}{*}{CLIP-L} 
     & CLIP-L &\textbf{61.9}&\textbf{70.0}&51.2&52.7&\textbf{59.0}&52.6 &58.4 &45.9 &43.7 &50.1\\ 
     & & BERT-B &56.9&66.0&45.9&49.6&54.6&56.5 &60.2 &44.6 &47.0 &52.1\\
     & & BERT-L &56.0&64.4&43.7&52.0&54.0&55.3 &60.5 &46.3 &46.3 &52.1\\

     \midrule
     \multirow{3}{*}{C} & \multirow{3}{*}{DINO-B} 
     & CLIP-B & 56.4&67.1&49.8&47.5&55.2&57.1 &64.6 &48.5 &46.2 &54.1\\
     & & BERT-B &56.7&65.1&48.5&45.2&53.9&54.2 &61.4 &48.7 &48.8 &53.3\\
     & & BERT-L &56.1&65.8&48.7&44.2&53.7&57.4 &65.1 &48.5 &46.3 &54.3\\
     
     \midrule
     \multirow{3}{*}{D} & \multirow{3}{*}{MoCo-B} 
     & CLIP-B & 58.9&67.0&47.7&51.7&56.3&56.4 &64.9 &47.1 &49.5 &54.5\\
     & & BERT-B &59.7&65.3&47.3&\textbf{53.4}&56.4&57.4 &66.1 &46.6 &51.2 &55.3\\
     & & BERT-L &59.1&65.7&49.5&53.1&56.8&54.9 &65.0 &49.8 &50.6 &55.0\\
     \bottomrule
    \end{tabular}}
    \caption{Ability of bridging different pre-trained backbones on PASCAL-5$^i$. 
    CLIP-B (or -L) means CLIP image/text encoders using ViT-B (or -L); 
    BERT-B (or -L) means BERT-Base (or -Large); 
    DINO-B or MoCo-B means using ViT-B backbone, respectively.}
    \label{table:backbone-ablation}
\end{table}

As illustrated in \tabref{table:backbone-ablation}, 
\textbf{early fusion} refers to our proposed fusion approach, 
while \textbf{late fusion} denotes similar the idea as LSeg~\cite{lseg}, 
where the visual or language features are separately processed with 6 MLPs layers,
and only fused at the last segmentation layer.

Here, we can make three observations:
(i) pairing the frozen visual and language models can be surprisingly powerful, 
even for models that are pre-trained on a corpus of uni-modal data, 
for example, in model-D, with MoCo-B as visual encoder, and BERT-L as language encoder;
(ii) early fusion consistently outperforms the late fusion, 
that validates the conjecture that visual-language correspondence can be better captured by allowing the feature of one modality to be updated in condition to the other;
(iii) the segmentation performance tends to improve with the model scale, 
for example, the model-B-CLIP works significantly better than model-A-CLIP, 
despite both are pre-trained with image-text pairs.
For latter experiments, we adopt the pair in model-B with both encoders CLIP-L for the best performance.

\subsection{Comparison with State-of-the-art}

Following LSeg~\cite{lseg}, 
we also compare our method with various zero-shot semantic segmentation methods: 
ZS3Net \cite{zs3net}, SPNet \cite{spnet} and LSeg \cite{lseg} on PASCAL-5$^i$ and COCO-20$^i$.

\begin{table}[!htb]
  \centering
  \setlength\tabcolsep{7pt}
  \resizebox{.95\textwidth}{!}{%
  \begin{tabular}{c|c|cccc|c|cccc|c}
    \toprule 
    \multirow{2}{*}{Model}&\multirow{2}{*}{Backbone}&\multicolumn{5}{c|}{PASCAL-$5^i$}&\multicolumn{5}{c}{COCO-$20^i$}\\
    \cmidrule(l){3-12}
    && $5^0$ & $5^1$ & $5^2$ & $5^3$ & mIoU & $20^0$ & $20^1$ & $20^2$ & $20^3$ & mIoU \\
    \midrule
    SPNet \cite{spnet}&ResNet101&23.8	&17.0&14.1&18.3&18.3&-&-&-&-&-\\
    ZS3Net \cite{zs3net}&ResNet101&40.8&39.4&39.3&33.6&38.3&18.8&20.1&24.8&20.5&21.1\\
    LSeg \cite{lseg}&ResNet101&52.8&53.8&44.4&38.5&47.4&22.1&25.1&24.9&21.5&23.4\\
    LSeg \cite{lseg}&ViT-L/16&61.3&63.6&43.1&41.0&52.3&28.1&27.5&30.0&23.2&27.2\\
    \midrule
    Fusioner~(ours) &ResNet101&46.8&56.0&42.2&40.7&46.4&26.7&34.1&26.3&23.4&27.6\\
    Fusioner~(ours) &ViT-L/14&\textbf{61.9}&\textbf{70.0}&\textbf{51.2}&\textbf{52.7}&\textbf{59.0}&\textbf{31.7}&\textbf{35.7}&\textbf{34.9}&\textbf{31.8}&\textbf{33.5}\\
    \bottomrule
    \end{tabular}}
  \caption{Comparison of mIoU on PASCAL-$5^i$ and COCO-$20^i$.}
  \label{table:benchmark}
\end{table}

As shown in \tabref{table:benchmark},
our proposed \textbf{Fusioner} with pre-trained frozen visual-language models~(CLIP-L) achieves state-of-the-art results on both PASCAL-5$^i$ and COCO-20$^i$, 
outperforming LSeg~\cite{lseg} by a significant margin on mIoU.
In contrast to PASCAL-5$^i$, COCO-20$^i$ is larger in scale and richer in objects,
for example, there may exist over 5 categories in one image, making it much more challenging. 
For late fusion models such as LSeg, 
the visual feature will be dominated by the most salient objects, 
however, our cross-modality fusion module can interchange information between visual and language features, adapt each other iteratively. This may explain the performance gap between us and LSeg.

\subsection{Robustness on Mosaic-4}

\begin{figure}[t]
  \centering
  \resizebox{\textwidth}{!}{
  \subfigure[Example image]{\
    \centering
    \includegraphics[width=.23\linewidth]{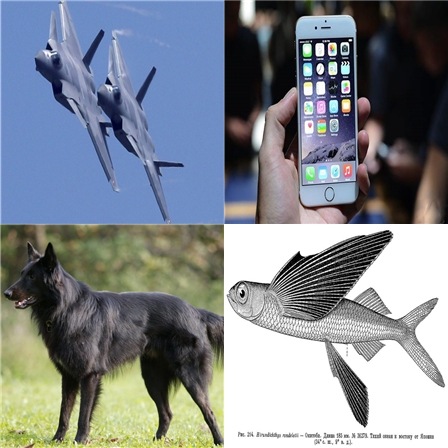}
    }
    \subfigure[Ground-truth]{\
    \centering
    \includegraphics[width=.23\linewidth]{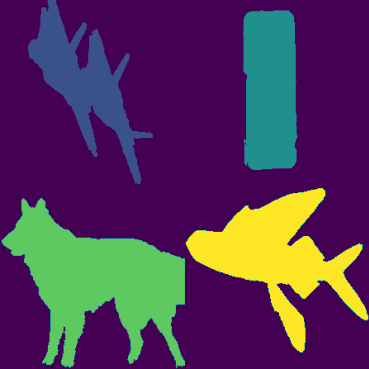}
    }
    \subfigure[Our prediction]{\
    \centering
    \includegraphics[width=.23\linewidth]{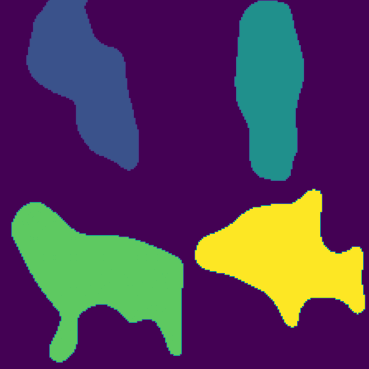}
    }
    \subfigure[LSeg prediction]{\
    \centering
    \includegraphics[width=.23\linewidth]{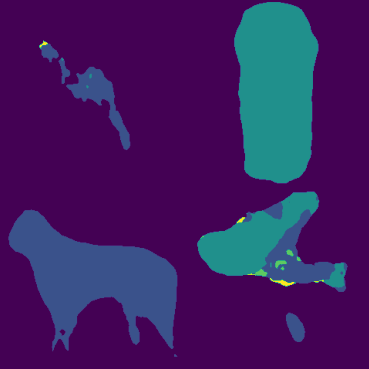}
  }}
  \vspace{-5pt}
  \caption{An example image and ground-truth of Mosaic-4 dataset, with models' predictions. The same color indicates the same category, \ie ``stealth\_aircraft'', ``iphone'', ``groenendael'' and ``abe's\_flyingfish'' from top-left to bottom-right. Our model can distinguish different categories inputs compared with LSeg. Best viewed in color.
  }
  \label{fig:fss4}
\end{figure}

Here, we measure the robustness of \textbf{Fusioner} trained on FSS-1000 using our synthesized Mosaic-4 dataset. 
We take $|\mathcal{W}|=4$, ~\ie, 4 category embeddings as input for one image, 
during training and testing, and generate 4 binary prediction for each category.
For LSeg, we input 4 categories together with text ``others'' representing the background, and conduct a pixel-wise classification into 5 categories.
As shown in \tabref{table:fss4}, our model can significantly outperform LSeg.
In \figref{fig:fss4}, 
our model can better differentiate different textual inputs along with their corresponding mask. 
However, LSeg is confused about ``groenendael'' and ``abe's\_flyingfish'', 
and segmenting ``stealth\_aircraft'' and ``iphone'' with large false positives.

\vspace{15pt}
\begin{minipage}{\textwidth}
\hspace{-10pt}
  \begin{minipage}[!t]{0.35\textwidth}
    \makeatletter\def\@captype{table}
    \centering
    \resizebox{.97\textwidth}{!}
    {
      \begin{tabular}{c|cc}
        \toprule
        Model & mIoU & FB-IoU \\\midrule
        LSeg \cite{lseg}  & 19.5 & 58.2   \\
        Fusioner (ours)  & \textbf{53.7} & \textbf{76.3}   \\\bottomrule
      \end{tabular}
    }
    \caption{Results on Mosaic-4.}
    \label{table:fss4}
  \end{minipage}
  \begin{minipage}[!t]{0.6\textwidth}
    \makeatletter\def\@captype{table}
    \centering
    \resizebox{0.97\textwidth}{!}
    {
      \begin{tabular}{ccc|cccc|c}
        \toprule
        Model &. Fusion     & Decoder      & $5^0$         & $5^1$         & $5^2$         & $5^3$         & mIoU          \\\midrule
        B0-CLIP-L & \Checkmark   & \Checkmark   & \textbf{61.9} & \textbf{70.0} & \textbf{51.2} & \textbf{52.7} & \textbf{59.0} \\
        B1-CLIP-L & \Checkmark   & \XSolidBrush & 59.4          & 63.7          & 47.3          & 44.0          & 53.6          \\
        B2-CLIP-L & \XSolidBrush & \Checkmark   & 48.3          & 54.4          & 41.4          & 42.2          & 46.6          \\
        B3-CLIP-L &  \XSolidBrush & \XSolidBrush & 16.7          & 21.7          & 20.0          & 20.9          & 19.8          \\
        \bottomrule
      \end{tabular}
    }
    \caption{Ablation study on PASCAL-5$^i$. }
    \label{table:components-ablation}
  \end{minipage}
    \hspace{\fill}
\end{minipage}

\subsection{Ablation Study}

To investigate the importance of each component in \textbf{Fusioner}, 
we conduct ablation studies on the cross-modality fusion and the visual decoder,
and change one variable at a time. 
All experiments are based on the best model in Table~\ref{table:backbone-ablation}, namely, model-B-CLIP-L.
As illustrated in \tabref{table:components-ablation}, 
model-B0-CLIP-L with both fusion and decoder gives the best results. 
Directly upsampling the visual feature without visual decoder leads to about a $5\%$ decline, 
while breaking the connection between visual and text modalities by skipping the fusion module results a sharp drop of 21\% in performance, 
which demonstrates the necessity of our cross-modality fusion module. 
However, when neither the fusion module nor the visual decoder is applied, 
no trainable parameters are introduce in the entire pipeline, which, unsurprisingly, 
yields the poorest result.

\subsection{Transferability to Other Datasets}
\label{transferability}

Ideally, open-vocabulary semantic segmentation should be able to handle any textual label regardless of the domain shift between different datasets. 
Here we evaluate on a more generalizable setting, 
that is, to test our COCO-trained model on PASCAL VOC following the work of \cite{boudiaf2021few}.
As shown in \tabref{table:transfer}, $20^i$ means the model was trained on fold $i$ of COCO-20$^i$ and tested on the whole PASCAL VOC dataset, after removing the seen classes in corresponding training split. Details can be found in supplementary. For evaluation, in addition to LSeg~\cite{lseg}, we also compare  the latest few-shot method RPMM~\cite{yang2020prototype}, CWT~\cite{lu2021simpler}, 
and PFENet~\cite{tian2020prior}. It can be seen from \tabref{table:transfer} that our method outperforms the previous state-of-the-art open-vocabulary method and is comparable to various few-shot methods.

\begin{figure}[!htp]
  \centering 
  \begin{minipage}[c]{.75\textwidth} 
    \makeatletter\def\@captype{table}
    \centering
    \resizebox{\textwidth}{!}
    {
      \begin{tabular}{ccc|cccc|c}
        \toprule
        Model&Backbone&Method&$20^0$&$20^1$&$20^2$&$20^3$&mIoU\\\midrule
        RPMM~\cite{yang2020prototype}&\multirow{3}{*}{ResNet50}&5-shot&40.2&58.0&55.2&61.8&53.8\\
        PFENet~\cite{tian2020prior}&&5-shot&45.1&\textbf{66.8}&\textbf{68.5}&\textbf{73.1}&63.4\\
        CWT~\cite{lu2021simpler}&&5-shot&\textbf{60.3}&65.8&67.1&72.8&\textbf{66.5}\\
        \midrule
        RPMM~\cite{yang2020prototype}&\multirow{3}{*}{ResNet50}&1-shot&36.3&55.0&52.5&54.6&49.6\\
        PFENet~\cite{tian2020prior} && 1-shot & 43.2 & \textbf{65.1}&\textbf{66.5}&\textbf{69.7}&\textbf{61.1}\\
        CWT~\cite{lu2021simpler}&&1-shot&\textbf{53.5}&59.2&60.2&64.9&59.5\\
        \midrule
        LSeg~\cite{lseg} & ResNet101&zero-shot&24.6& - &34.7&35.9 & 31.7 \tnote{*}\\
        Fusioner (ours) & ResNet101&zero-shot&31.0&53.7&41.7&51.3&44.4\\
        Fusioner (ours) & ViT-L&zero-shot&\textbf{39.9}&\textbf{70.7}&\textbf{47.8}&\textbf{67.6}&\textbf{56.5}\\
        \bottomrule
        \end{tabular}
    }
  \end{minipage} 
  \begin{minipage}[c]{0.24\textwidth} 
    \makeatletter\def\@captype{table}
    \centering 
    \caption{Transferability from COCO-20$^i$ to PASCAL VOC. 
    Here, LSeg results are generated by averaging the three officially released checkpoints (no ViT-L backbone, only fold 0,2,3 for ResNet).} 
    \label{table:transfer} 
  \end{minipage}%
\end{figure}

\subsection{Qualitative Results}

\begin{figure}[!htp]

\setlength{\subfigcapskip}{-3pt}
\setlength{\subfigtopskip}{2pt}
  \setcounter{subfigure}{0}
  \subfigure[image]{
      \centering
      \includegraphics[width=.19\linewidth]{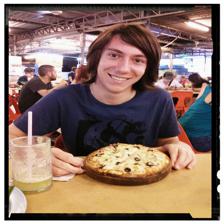}
  }
  \hspace{-7pt}
  \subfigure[pizza*]{
      \centering
      \includegraphics[width=.19\linewidth]{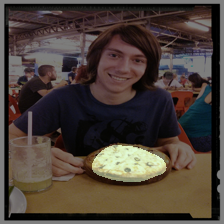}
  }
  \hspace{-7pt}
  \subfigure[cup*]{
      \centering
      \includegraphics[width=.19\linewidth]{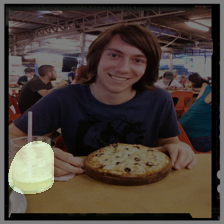}
  }
  \hspace{-7pt}
  \subfigure[diningtable]{
      \centering
      \includegraphics[width=.19\linewidth]{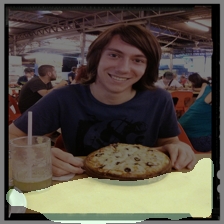}
  }
  \hspace{-7pt}
  \subfigure[person]{
      \centering
      \includegraphics[width=.19\linewidth]{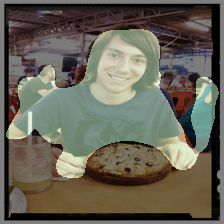}
  }

 \vspace{-5pt} 
  \setcounter{subfigure}{0}
  \subfigure[image]{
      \centering
      \includegraphics[width=.19\linewidth]{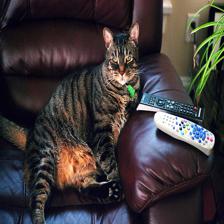}
  }
  \hspace{-7pt}
  \subfigure[remote*]{
      \centering
      \includegraphics[width=.19\linewidth]{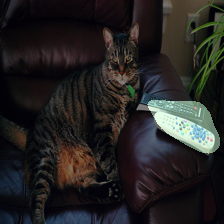}
  }
  \hspace{-7pt}
  \subfigure[sofa*]{
      \centering
      \includegraphics[width=.19\linewidth]{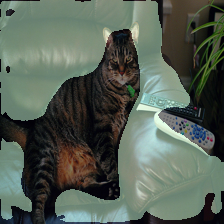}
  }
  \hspace{-7pt}
  \subfigure[cat]{
      \centering
      \includegraphics[width=.19\linewidth]{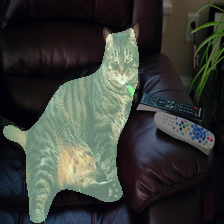}
  }
  \hspace{-7pt}
  \subfigure[pottedplant]{
      \centering
      \includegraphics[width=.19\linewidth]{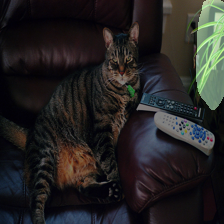}
  }

  \vspace{-5pt}
  \setcounter{subfigure}{0}
  \subfigure[image]{
      \centering
      \includegraphics[width=.19\linewidth]{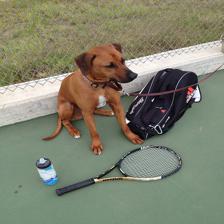}
  }
  \hspace{-7pt}
  \subfigure[tennisracket*]{
      \centering
      \includegraphics[width=.19\linewidth]{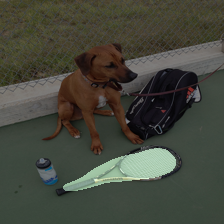}
  }
  \hspace{-7pt}
  \subfigure[handbag*]{
      \centering
      \includegraphics[width=.19\linewidth]{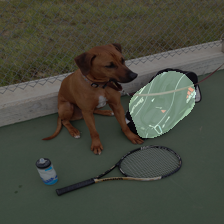}
  }
  \hspace{-7pt}
  \subfigure[dog]{
      \centering
      \includegraphics[width=.19\linewidth]{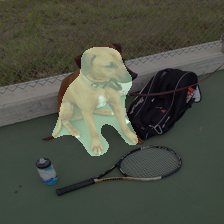}
  }
  \hspace{-7pt}
  \subfigure[bottle]{
      \centering
      \includegraphics[width=.19\linewidth]{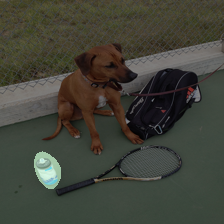}
  }
  \vspace{-5pt}
  \caption{Qualitative results on COCO-20$^i$. (a) input images, (b)-(e) segmentation masks for different categories. Unseen categories are marked as *.
  }
  \label{fig:visualize}
\end{figure}

In \figref{fig:visualize}, we show the qualitative results for our model on open-vocabulary segmentation. Specifically, the subimages in each row include the original image, 
and segmentation results of seen and unseen (marked as *) categories. 
As can be seen, our proposed \textbf{Fusioner} can successfully predict the unseen categories.

\section{Conclusion}

With the growing interest in Foundation Models~\cite{bommasani2021opportunities}, 
we believe it will be of great significance for the community, 
to efficiently adapt these powerful vision and language models for the downstream task of interest. Here, we introduce \textbf{Fusioner}, 
a simple, lightweight cross-modality fusion module, 
that explicitly bridged a variety of self-supervised pre-trained visual/language models for open-vocabulary semantic segmentation. 
We evaluate on two standard benchmarks~(PASCAL and COCO), and conduct thorough ablation studies to demonstrate the effectiveness of our model.
Despite the simplicity of the proposed idea, 
we demonstrate state-of-the-art on all standard benchmarks.

\section*{Acknowledgments}

This work is supported by the National Key R\&D Program of China (No. 2019YFB1804304), 111 plan (No. BP0719010),  and STCSM (No. 18DZ2270700, No. 21DZ1100100), and State Key Laboratory of UHD Video and Audio Production and Presentation.

\bibliography{egbib}

\begin{thebibliography}{54}
\providecommand{\natexlab}[1]{#1}
\providecommand{\url}[1]{\texttt{#1}}
\expandafter\ifx\csname urlstyle\endcsname\relax
  \providecommand{\doi}[1]{doi: #1}\else
  \providecommand{\doi}{doi: \begingroup \urlstyle{rm}\Url}\fi

\bibitem[Alayrac et~al.(2022)Alayrac, Donahue, Luc, Miech, Barr, Hasson, Lenc,
  Mensch, Millican, Reynolds, et~al.]{flamingo}
Jean-Baptiste Alayrac, Jeff Donahue, Pauline Luc, Antoine Miech, Iain Barr,
  Yana Hasson, Karel Lenc, Arthur Mensch, Katie Millican, Malcolm Reynolds,
  et~al.
\newblock Flamingo: a visual language model for few-shot learning.
\newblock \emph{arXiv preprint arXiv:2204.14198}, 2022.

\bibitem[Baek et~al.(2021)Baek, Oh, and Ham]{baek2021exploiting}
Donghyeon Baek, Youngmin Oh, and Bumsub Ham.
\newblock Exploiting a joint embedding space for generalized zero-shot semantic
  segmentation.
\newblock In \emph{Proceedings of the IEEE/CVF International Conference on
  Computer Vision}, pages 9536--9545, 2021.

\bibitem[Bommasani and et~al.(2021)]{bommasani2021opportunities}
Rishi Bommasani and et~al.
\newblock On the opportunities and risks of foundation models.
\newblock \emph{arXiv preprint arXiv:2108.07258}, 2021.

\bibitem[Boudiaf et~al.(2021)Boudiaf, Kervadec, Masud, Piantanida, Ben~Ayed,
  and Dolz]{boudiaf2021few}
Malik Boudiaf, Hoel Kervadec, Ziko~Imtiaz Masud, Pablo Piantanida, Ismail
  Ben~Ayed, and Jose Dolz.
\newblock Few-shot segmentation without meta-learning: A good transductive
  inference is all you need?
\newblock In \emph{Proceedings of the IEEE/CVF Conference on Computer Vision
  and Pattern Recognition}, pages 13979--13988, 2021.

\bibitem[Brown et~al.(2020)Brown, Mann, Ryder, Subbiah, Kaplan, Dhariwal,
  Neelakantan, Shyam, Sastry, Askell, et~al.]{gpt3}
Tom Brown, Benjamin Mann, Nick Ryder, Melanie Subbiah, Jared~D Kaplan, Prafulla
  Dhariwal, Arvind Neelakantan, Pranav Shyam, Girish Sastry, Amanda Askell,
  et~al.
\newblock Language models are few-shot learners.
\newblock \emph{Advances in Neural Information Processing Systems},
  33:\penalty0 1877--1901, 2020.

\bibitem[Bucher et~al.(2019)Bucher, Vu, Cord, and P{\'e}rez]{zs3net}
Maxime Bucher, Tuan-Hung Vu, Matthieu Cord, and Patrick P{\'e}rez.
\newblock Zero-shot semantic segmentation.
\newblock \emph{Advances in Neural Information Processing Systems}, 32, 2019.

\bibitem[Caron et~al.(2020)Caron, Misra, Mairal, Goyal, Bojanowski, and
  Joulin]{swav}
Mathilde Caron, Ishan Misra, Julien Mairal, Priya Goyal, Piotr Bojanowski, and
  Armand Joulin.
\newblock Unsupervised learning of visual features by contrasting cluster
  assignments.
\newblock \emph{Advances in Neural Information Processing Systems},
  33:\penalty0 9912--9924, 2020.

\bibitem[Caron et~al.(2021)Caron, Touvron, Misra, J{\'e}gou, Mairal,
  Bojanowski, and Joulin]{dino}
Mathilde Caron, Hugo Touvron, Ishan Misra, Herv{\'e} J{\'e}gou, Julien Mairal,
  Piotr Bojanowski, and Armand Joulin.
\newblock Emerging properties in self-supervised vision transformers.
\newblock In \emph{Proceedings of the IEEE/CVF International Conference on
  Computer Vision}, pages 9650--9660, 2021.

\bibitem[Chen et~al.(2017)Chen, Papandreou, Kokkinos, Murphy, and
  Yuille]{DeepLab}
Liang-Chieh Chen, George Papandreou, Iasonas Kokkinos, Kevin Murphy, and Alan~L
  Yuille.
\newblock Deeplab: Semantic image segmentation with deep convolutional nets,
  atrous convolution, and fully connected crfs.
\newblock \emph{IEEE Transactions on Pattern Analysis and Machine
  Intelligence}, 40\penalty0 (4):\penalty0 834--848, 2017.

\bibitem[Chen et~al.(2020{\natexlab{a}})Chen, Kornblith, Norouzi, and
  Hinton]{simclr}
Ting Chen, Simon Kornblith, Mohammad Norouzi, and Geoffrey Hinton.
\newblock A simple framework for contrastive learning of visual
  representations.
\newblock In \emph{International Conference on Machine Learning}, pages
  1597--1607. PMLR, 2020{\natexlab{a}}.

\bibitem[Chen and He(2021)]{simsiam}
Xinlei Chen and Kaiming He.
\newblock Exploring simple siamese representation learning.
\newblock In \emph{Proceedings of the IEEE/CVF Conference on Computer Vision
  and Pattern Recognition}, pages 15750--15758, 2021.

\bibitem[Chen et~al.(2021)Chen, Xie, and He]{mocov3}
Xinlei Chen, Saining Xie, and Kaiming He.
\newblock An empirical study of training self-supervised vision transformers.
\newblock In \emph{Proceedings of the IEEE/CVF International Conference on
  Computer Vision}, pages 9640--9649, 2021.

\bibitem[Chen et~al.(2020{\natexlab{b}})Chen, Li, Yu, El~Kholy, Ahmed, Gan,
  Cheng, and Liu]{chen2020uniter}
Yen-Chun Chen, Linjie Li, Licheng Yu, Ahmed El~Kholy, Faisal Ahmed, Zhe Gan,
  Yu~Cheng, and Jingjing Liu.
\newblock Uniter: Universal image-text representation learning.
\newblock In \emph{European Conference on Computer Vision}, pages 104--120.
  Springer, 2020{\natexlab{b}}.

\bibitem[Chowdhery and et~al.(2022)]{chowdhery2022palm}
Aakanksha Chowdhery and et~al.
\newblock Palm: Scaling language modeling with pathways.
\newblock \emph{arXiv preprint arXiv:2204.02311}, 2022.

\bibitem[Dosovitskiy et~al.(2021)Dosovitskiy, Beyer, Kolesnikov, Weissenborn,
  Zhai, Unterthiner, Dehghani, Minderer, Heigold, Gelly, Uszkoreit, and
  Houlsby]{vit}
Alexey Dosovitskiy, Lucas Beyer, Alexander Kolesnikov, Dirk Weissenborn,
  Xiaohua Zhai, Thomas Unterthiner, Mostafa Dehghani, Matthias Minderer, Georg
  Heigold, Sylvain Gelly, Jakob Uszkoreit, and Neil Houlsby.
\newblock An image is worth 16x16 words: Transformers for image recognition at
  scale.
\newblock In \emph{International Conference on Learning Representations}, 2021.

\bibitem[Everingham et~al.(2010)Everingham, Van~Gool, Williams, Winn, and
  Zisserman]{everingham2010pascal}
Mark Everingham, Luc Van~Gool, Christopher~KI Williams, John Winn, and Andrew
  Zisserman.
\newblock The pascal visual object classes (voc) challenge.
\newblock \emph{International journal of computer vision}, 88\penalty0
  (2):\penalty0 303--338, 2010.

\bibitem[Feng et~al.(2022)Feng, Zhong, Jie, Chu, Ren, Wei, Xie, and
  Ma]{feng2022promptdet}
Chengjian Feng, Yujie Zhong, Zequn Jie, Xiangxiang Chu, Haibing Ren, Xiaolin
  Wei, Weidi Xie, and Lin Ma.
\newblock Promptdet: Expand your detector vocabulary with uncurated images.
\newblock In \emph{European Conference on Computer Vision}, 2022.

\bibitem[Grill et~al.(2020)Grill, Strub, Altch{\'e}, Tallec, Richemond,
  Buchatskaya, Doersch, Avila~Pires, Guo, Gheshlaghi~Azar, et~al.]{byol}
Jean-Bastien Grill, Florian Strub, Florent Altch{\'e}, Corentin Tallec, Pierre
  Richemond, Elena Buchatskaya, Carl Doersch, Bernardo Avila~Pires, Zhaohan
  Guo, Mohammad Gheshlaghi~Azar, et~al.
\newblock Bootstrap your own latent-a new approach to self-supervised learning.
\newblock \emph{Advances in Neural Information Processing Systems},
  33:\penalty0 21271--21284, 2020.

\bibitem[Gu et~al.(2022{\natexlab{a}})Gu, Lin, Kuo, and Cui]{vild}
Xiuye Gu, Tsung-Yi Lin, Weicheng Kuo, and Yin Cui.
\newblock Open-vocabulary object detection via vision and language knowledge
  distillation.
\newblock In \emph{International Conference on Learning Representations},
  2022{\natexlab{a}}.

\bibitem[Gu et~al.(2020)Gu, Zhou, Niu, Zhao, and Zhang]{gu2020context}
Zhangxuan Gu, Siyuan Zhou, Li~Niu, Zihan Zhao, and Liqing Zhang.
\newblock Context-aware feature generation for zero-shot semantic segmentation.
\newblock In \emph{Proceedings of the 28th ACM International Conference on
  Multimedia}, pages 1921--1929, 2020.

\bibitem[Gu et~al.(2022{\natexlab{b}})Gu, Zhou, Niu, Zhao, and
  Zhang]{gu2022pixel}
Zhangxuan Gu, Siyuan Zhou, Li~Niu, Zihan Zhao, and Liqing Zhang.
\newblock From pixel to patch: Synthesize context-aware features for zero-shot
  semantic segmentation.
\newblock \emph{IEEE Transactions on Neural Networks and Learning Systems},
  2022{\natexlab{b}}.

\bibitem[Hariharan et~al.(2011)Hariharan, Arbel{\'a}ez, Bourdev, Maji, and
  Malik]{hariharan2011semantic}
Bharath Hariharan, Pablo Arbel{\'a}ez, Lubomir Bourdev, Subhransu Maji, and
  Jitendra Malik.
\newblock Semantic contours from inverse detectors.
\newblock In \emph{Proceedings of the IEEE/CVF International Conference on
  Computer Vision}, pages 991--998. IEEE, 2011.

\bibitem[Hessel et~al.(2021)Hessel, Holtzman, Forbes, Le~Bras, and
  Choi]{hessel2021clipscore}
Jack Hessel, Ari Holtzman, Maxwell Forbes, Ronan Le~Bras, and Yejin Choi.
\newblock Clipscore: A reference-free evaluation metric for image captioning.
\newblock In \emph{Proceedings of the 2021 Conference on Empirical Methods in
  Natural Language Processing}, pages 7514--7528, 2021.

\bibitem[Jia et~al.(2021)Jia, Yang, Xia, Chen, Parekh, Pham, Le, Sung, Li, and
  Duerig]{align}
Chao Jia, Yinfei Yang, Ye~Xia, Yi-Ting Chen, Zarana Parekh, Hieu Pham, Quoc Le,
  Yun-Hsuan Sung, Zhen Li, and Tom Duerig.
\newblock Scaling up visual and vision-language representation learning with
  noisy text supervision.
\newblock In \emph{International Conference on Machine Learning}, pages
  4904--4916. PMLR, 2021.

\bibitem[Kenton and Toutanova(2019)]{bert}
Jacob Devlin Ming-Wei~Chang Kenton and Lee~Kristina Toutanova.
\newblock Bert: Pre-training of deep bidirectional transformers for language
  understanding.
\newblock In \emph{Proceedings of NAACL-HLT}, pages 4171--4186, 2019.

\bibitem[Lei et~al.(2021)Lei, Li, Zhou, Gan, Berg, Bansal, and
  Liu]{lei2021less}
Jie Lei, Linjie Li, Luowei Zhou, Zhe Gan, Tamara~L Berg, Mohit Bansal, and
  Jingjing Liu.
\newblock Less is more: Clipbert for video-and-language learning via sparse
  sampling.
\newblock In \emph{Proceedings of the IEEE/CVF Conference on Computer Vision
  and Pattern Recognition}, pages 7331--7341, 2021.

\bibitem[Li et~al.(2022)Li, Weinberger, Belongie, Koltun, and Ranftl]{lseg}
Boyi Li, Kilian~Q Weinberger, Serge Belongie, Vladlen Koltun, and Rene Ranftl.
\newblock Language-driven semantic segmentation.
\newblock In \emph{International Conference on Learning Representations}, 2022.

\bibitem[Li et~al.(2020{\natexlab{a}})Li, Duan, Fang, Gong, and
  Jiang]{li2020unicoder}
Gen Li, Nan Duan, Yuejian Fang, Ming Gong, and Daxin Jiang.
\newblock Unicoder-vl: A universal encoder for vision and language by
  cross-modal pre-training.
\newblock In \emph{Proceedings of the AAAI Conference on Artificial
  Intelligence}, pages 11336--11344, 2020{\natexlab{a}}.

\bibitem[Li et~al.(2020{\natexlab{b}})Li, Wei, and Yang]{csrl}
Peike Li, Yunchao Wei, and Yi~Yang.
\newblock Consistent structural relation learning for zero-shot segmentation.
\newblock \emph{Advances in Neural Information Processing Systems},
  33:\penalty0 10317--10327, 2020{\natexlab{b}}.

\bibitem[Li et~al.(2020{\natexlab{c}})Li, Yin, Li, Zhang, Hu, Zhang, Wang, Hu,
  Dong, Wei, et~al.]{li2020oscar}
Xiujun Li, Xi~Yin, Chunyuan Li, Pengchuan Zhang, Xiaowei Hu, Lei Zhang, Lijuan
  Wang, Houdong Hu, Li~Dong, Furu Wei, et~al.
\newblock Oscar: Object-semantics aligned pre-training for vision-language
  tasks.
\newblock In \emph{European Conference on Computer Vision}, pages 121--137.
  Springer, 2020{\natexlab{c}}.

\bibitem[Lin et~al.(2014)Lin, Maire, Belongie, Hays, Perona, Ramanan,
  Doll{\'a}r, and Zitnick]{lin2014microsoft}
Tsung-Yi Lin, Michael Maire, Serge Belongie, James Hays, Pietro Perona, Deva
  Ramanan, Piotr Doll{\'a}r, and C~Lawrence Zitnick.
\newblock Microsoft coco: Common objects in context.
\newblock In \emph{European Conference on Computer Vision}, pages 740--755.
  Springer, 2014.

\bibitem[Lu et~al.(2019)Lu, Batra, Parikh, and Lee]{lu2019vilbert}
Jiasen Lu, Dhruv Batra, Devi Parikh, and Stefan Lee.
\newblock Vilbert: Pretraining task-agnostic visiolinguistic representations
  for vision-and-language tasks.
\newblock \emph{Advances in Neural Information Processing Systems}, 32, 2019.

\bibitem[Lu et~al.(2021)Lu, He, Zhu, Zhang, Song, and Xiang]{lu2021simpler}
Zhihe Lu, Sen He, Xiatian Zhu, Li~Zhang, Yi-Zhe Song, and Tao Xiang.
\newblock Simpler is better: Few-shot semantic segmentation with classifier
  weight transformer.
\newblock In \emph{Proceedings of the IEEE/CVF International Conference on
  Computer Vision}, pages 8741--8750, 2021.

\bibitem[Melas-Kyriazi et~al.(2022)Melas-Kyriazi, Laina, Rupprecht, and
  Vedaldi]{Melaskyriazi22}
Luke Melas-Kyriazi, Iro Laina, Christian Rupprecht, and Andrea Vedaldi.
\newblock Deep spectral methods: A surprisingly strong baseline for
  unsupervised semantic segmentation and localization.
\newblock In \emph{Proceedings of the IEEE/CVF Conference on Computer Vision
  and Pattern Recognition}, 2022.

\bibitem[Patashnik et~al.(2021)Patashnik, Wu, Shechtman, Cohen-Or, and
  Lischinski]{patashnik2021styleclip}
Or~Patashnik, Zongze Wu, Eli Shechtman, Daniel Cohen-Or, and Dani Lischinski.
\newblock Styleclip: Text-driven manipulation of stylegan imagery.
\newblock In \emph{Proceedings of the IEEE/CVF International Conference on
  Computer Vision}, pages 2085--2094, 2021.

\bibitem[Radford et~al.(2018)Radford, Narasimhan, Salimans, Sutskever,
  et~al.]{gpt1}
Alec Radford, Karthik Narasimhan, Tim Salimans, Ilya Sutskever, et~al.
\newblock Improving language understanding by generative pre-training.
\newblock \emph{OpenAI blog}, 2018.

\bibitem[Radford et~al.(2019)Radford, Wu, Child, Luan, Amodei, Sutskever,
  et~al.]{gpt2}
Alec Radford, Jeffrey Wu, Rewon Child, David Luan, Dario Amodei, Ilya
  Sutskever, et~al.
\newblock Language models are unsupervised multitask learners.
\newblock \emph{OpenAI blog}, 2019.

\bibitem[Radford et~al.(2021)Radford, Kim, Hallacy, Ramesh, Goh, Agarwal,
  Sastry, Askell, Mishkin, Clark, et~al.]{clip}
Alec Radford, Jong~Wook Kim, Chris Hallacy, Aditya Ramesh, Gabriel Goh,
  Sandhini Agarwal, Girish Sastry, Amanda Askell, Pamela Mishkin, Jack Clark,
  et~al.
\newblock Learning transferable visual models from natural language
  supervision.
\newblock In \emph{International Conference on Machine Learning}, pages
  8748--8763. PMLR, 2021.

\bibitem[Raffel et~al.(2020)Raffel, Shazeer, Roberts, Lee, Narang, Matena,
  Zhou, Li, and Liu]{t5}
Colin Raffel, Noam Shazeer, Adam Roberts, Katherine Lee, Sharan Narang, Michael
  Matena, Yanqi Zhou, Wei Li, and Peter~J. Liu.
\newblock Exploring the limits of transfer learning with a unified text-to-text
  transformer.
\newblock \emph{Journal of Machine Learning Research}, 21\penalty0
  (140):\penalty0 1--67, 2020.

\bibitem[Ranftl et~al.(2021)Ranftl, Bochkovskiy, and Koltun]{DPT}
Ren{\'e} Ranftl, Alexey Bochkovskiy, and Vladlen Koltun.
\newblock Vision transformers for dense prediction.
\newblock In \emph{Proceedings of the IEEE/CVF International Conference on
  Computer Vision}, pages 12159--12168, 2021.

\bibitem[Rao et~al.(2022)Rao, Zhao, Chen, Tang, Zhu, Huang, Zhou, and
  Lu]{rao2022denseclip}
Yongming Rao, Wenliang Zhao, Guangyi Chen, Yansong Tang, Zheng Zhu, Guan Huang,
  Jie Zhou, and Jiwen Lu.
\newblock Denseclip: Language-guided dense prediction with context-aware
  prompting.
\newblock In \emph{Proceedings of the IEEE/CVF Conference on Computer Vision
  and Pattern Recognition}, pages 18082--18091, 2022.

\bibitem[Ronneberger et~al.(2015)Ronneberger, Fischer, and Brox]{UNet}
Olaf Ronneberger, Philipp Fischer, and Thomas Brox.
\newblock U-net: Convolutional networks for biomedical image segmentation.
\newblock In \emph{Medical Image Computing and Computer-Assisted Intervention},
  pages 234--241, 2015.

\bibitem[Shelhamer et~al.(2017)Shelhamer, Long, and Darrell]{FCN}
Evan Shelhamer, Jonathan Long, and Trevor Darrell.
\newblock Fully convolutional networks for semantic segmentation.
\newblock \emph{IEEE Transactions on Pattern Analysis and Machine
  Intelligence}, 39:\penalty0 640--651, 2017.

\bibitem[Shin et~al.(2022)Shin, Albanie, and Xie]{Shin22}
Gyungin Shin, Samuel Albanie, and Weidi Xie.
\newblock Unsupervised salient object detection with spectral cluster voting.
\newblock In \emph{IEEE Conference on Computer Vision and Pattern Recognition
  L3DIVU Workshop}, 2022.

\bibitem[Su et~al.(2020)Su, Zhu, Cao, Li, Lu, Wei, and Dai]{su2019vl}
Weijie Su, Xizhou Zhu, Yue Cao, Bin Li, Lewei Lu, Furu Wei, and Jifeng Dai.
\newblock Vl-bert: Pre-training of generic visual-linguistic representations.
\newblock In \emph{International Conference on Learning Representations}, 2020.

\bibitem[Tian et~al.(2020)Tian, Zhao, Shu, Yang, Li, and Jia]{tian2020prior}
Zhuotao Tian, Hengshuang Zhao, Michelle Shu, Zhicheng Yang, Ruiyu Li, and Jiaya
  Jia.
\newblock Prior guided feature enrichment network for few-shot segmentation.
\newblock \emph{IEEE Transactions on Pattern Analysis and Machine
  Intelligence}, 2020.

\bibitem[Tsimpoukelli et~al.(2021)Tsimpoukelli, Menick, Cabi, Eslami, Vinyals,
  and Hill]{frozen}
Maria Tsimpoukelli, Jacob~L Menick, Serkan Cabi, SM~Eslami, Oriol Vinyals, and
  Felix Hill.
\newblock Multimodal few-shot learning with frozen language models.
\newblock \emph{Advances in Neural Information Processing Systems},
  34:\penalty0 200--212, 2021.

\bibitem[Wang et~al.(2022{\natexlab{a}})Wang, Shen, Hu, Yuan, Crowley, and
  Vaufreydaz]{Wang22tokencut}
Yangtao Wang, Xi~Shen, Shell~Xu Hu, Yuan Yuan, James~L. Crowley, and Dominique
  Vaufreydaz.
\newblock Self-supervised transformers for unsupervised object discovery using
  normalized cut.
\newblock In \emph{Proceedings of the IEEE/CVF Conference on Computer Vision
  and Pattern Recognition}, 2022{\natexlab{a}}.

\bibitem[Wang et~al.(2022{\natexlab{b}})Wang, Lu, Li, Tao, Guo, Gong, and
  Liu]{wang2022cris}
Zhaoqing Wang, Yu~Lu, Qiang Li, Xunqiang Tao, Yandong Guo, Mingming Gong, and
  Tongliang Liu.
\newblock Cris: Clip-driven referring image segmentation.
\newblock In \emph{Proceedings of the IEEE/CVF Conference on Computer Vision
  and Pattern Recognition}, pages 11686--11695, 2022{\natexlab{b}}.

\bibitem[Wei et~al.(2020)Wei, Li, Chen, Tai, and Tang]{fss1000}
Tianhan Wei, Xiang Li, Yau~Pun Chen, Yu-Wing Tai, and Chi-Keung Tang.
\newblock Fss-1000: A 1000-class dataset for few-shot segmentation.
\newblock In \emph{Proceedings of the IEEE/CVF Conference on Computer Vision
  and Pattern Recognition}, pages 2866--2875, 2020.

\bibitem[Xian et~al.(2019)Xian, Choudhury, He, Schiele, and Akata]{spnet}
Yongqin Xian, Subhabrata Choudhury, Yang He, Bernt Schiele, and Zeynep Akata.
\newblock Semantic projection network for zero-and few-label semantic
  segmentation.
\newblock In \emph{Proceedings of the IEEE/CVF Conference on Computer Vision
  and Pattern Recognition}, pages 8256--8265, 2019.

\bibitem[Yang et~al.(2020)Yang, Liu, Li, Jiao, and Ye]{yang2020prototype}
Boyu Yang, Chang Liu, Bohao Li, Jianbin Jiao, and Qixiang Ye.
\newblock Prototype mixture models for few-shot semantic segmentation.
\newblock In \emph{European Conference on Computer Vision}, pages 763--778.
  Springer, 2020.

\bibitem[Yuan et~al.(2021)Yuan, Chen, Chen, Codella, Dai, Gao, Hu, Huang, Li,
  Li, et~al.]{yuan2021florence}
Lu~Yuan, Dongdong Chen, Yi-Ling Chen, Noel Codella, Xiyang Dai, Jianfeng Gao,
  Houdong Hu, Xuedong Huang, Boxin Li, Chunyuan Li, et~al.
\newblock Florence: A new foundation model for computer vision.
\newblock \emph{arXiv preprint arXiv:2111.11432}, 2021.

\bibitem[Zeng et~al.(2022)Zeng, Wong, Welker, Choromanski, Tombari, Purohit,
  Ryoo, Sindhwani, Lee, Vanhoucke, et~al.]{socratic}
Andy Zeng, Adrian Wong, Stefan Welker, Krzysztof Choromanski, Federico Tombari,
  Aveek Purohit, Michael Ryoo, Vikas Sindhwani, Johnny Lee, Vincent Vanhoucke,
  et~al.
\newblock Socratic models: Composing zero-shot multimodal reasoning with
  language.
\newblock \emph{arXiv preprint arXiv:2204.00598}, 2022.

\end{thebibliography}

\section{Appendix}

In this supplementary material, 
we start by giving a detailed description about our experimental setup, 
including datasets split, and prompt templates. 
The results for ablation study on the architecture of cross-modality fusion module are then presented. 
Lastly, more qualitative visualization results are displayed.

\subsection{Datasets Settings}

In \tabref{table:dataset}, we provide the detailed datasets split settings used in our experiments. 
Datasets PASCAL-5$^i$~\cite{everingham2010pascal,hariharan2011semantic} and COCO-20$^i$~\cite{lin2014microsoft} follow the split settings proposed in LSeg \cite{lseg}. 
COCO-20$^i\to$PASCAL VOC refers to the transferability experiment mentioned in Section 4.6 of the main paper.

\begin{table}[!htb]
    \centering
    \setlength\tabcolsep{2pt}
    \resizebox{.96\textwidth}{!}{%
        \begin{tabular}{c|cccc}
            \toprule
            Dataset&fold 0&fold 1&fold 2&fold 3\\
            \midrule
            PASCAL-$5^i$&\makecell{aeroplane, bicycle, \\bird, boat, bottle}&\makecell{bus, car, cat, \\chair, cow}&\makecell{diningtable, dog, horse, \\motorbike, person}&\makecell{pottedplant, sheep, \\sofa, train, tv/monitor}\\\midrule
            COCO-$20^i$&\makecell{person, aeroplane, boat, \\parkingmeter, dog, elephant, \\backpack, suitcase, sportsball, \\skateboard, wineglass, spoon, \\sandwich, hotdog, chair, \\diningtable, mouse, microwave, \\refrigerator, scissors}&\makecell{bicycle, bus, trafficlight, \\bench, horse, bear, \\umbrella, frisbee, kite, \\surfboard, cup, bowl, \\orange, pizza, sofa, \\toilet, remote, oven, \\book, teddybear}&\makecell{car, train, firehydrant, \\bird, sheep, zebra, \\handbag, skis, baseballbat, \\tennisracket, fork, banana, \\broccoli, donut, pottedplant, \\tvmonitor, keyboard, toaster, \\clock, hairdrier}&\makecell{motorbike, truck, stopsign, \\cat, cow, giraffe, tie, \\snowboard, baseballglove, \\bottle, knife, apple, \\carrot, cake, bed, \\laptop, cellphone, sink, \\vase, toothbrush}\\\midrule
            \makecell{COCO-$20^i\to$\\PASCAL VOC}&\makecell{aeroplane, boat, chair, \\diningtable, dog, person}&\makecell{ bicycle, bus, \\horse, sofa}&\makecell{ bird, car, pottedplant, \\sheep, train, tvmonitor}&\makecell{ bottle, cat, \\cow, motorbike}\\
            \bottomrule
        \end{tabular}}
    \caption{Data split for PASCAL-5$^i$, COCO-$20^i$, and COCO-$20^i\to$PASCAL VOC. PASCAL-5$^i$ and COCO-$20^i$ use 4-fold cross-validation, and categories in each row are for test and the rest classes are used for training. In COCO-20$^i\to$PASCAL VOC, categories in fold $i$ are novel classes in PASCAL VOC after removing the seen classes in corresponding training split on fold $i$ of COCO-20$^i$.}
    \label{table:dataset}
\end{table}

\clearpage
\subsection{Prompt Templates}

Before feeding the semantic category into text encoder, we use multiple prompt templates as decorations to generate text embeddings for one category, and then ensemble these embeddings by averging.
The following are the prompt templates we used:

\begin{figure}[!htb]
\vspace{15pt}
    \begin{small}
	\begin{verbatim}
    'a bad photo of the {category}.',
    'a photo of the large {category}.',
    'a photo of the small {category}.',
    'a cropped photo of a {category}.',
    'This is a photo of a {category}',
    'This is a photo of a small {category}',
    'This is a photo of a medium {category}',
    'This is a photo of a large {category}',
    'This is a masked photo of a {category}',
    'This is a masked photo of a small {category}',
    'This is a masked photo of a medium {category}',
    'This is a masked photo of a large {category}',
    'This is a cropped photo of a {category}',
    'This is a cropped photo of a small {category}',
    'This is a cropped photo of a medium {category}',
    'This is a cropped photo of a large {category}',
    'A photo of a {category} in the scene',
    'a bad photo of the {category} in the scene',
    'a photo of the large {category} in the scene',
    'a photo of the small {category} in the scene',
    'a cropped photo of a {category} in the scene',
    'a photo of a masked {category} in the scene',
    'There is a {category} in the scene',
    'There is the {category} in the scene',
    'This is a {category} in the scene',
    'This is the {category} in the scene',
    'This is one {category} in the scene',
    'There is a masked {category} in the scene',
    'There is the masked {category} in the scene',
    'This is a masked {category} in the scene',
    'This is the masked {category} in the scene',
    'This is one masked {category} in the scene',
	\end{verbatim}
\end{small}
\end{figure}

\clearpage

\subsection{Ablations of Cross-modality Fusion Module}

We conduct an ablation study on the architecture of the cross-modality fusion module, \ie, the number of layers, heads, and width of the transformer. The $\text{width}/\text{head}$ ratio is set to $64$ in this ablation.
In \tabref{table:architecture-ablation}, we can see that, a lightweight fusion module (1-layer transformer) already works reasonably well, 
and a steady improvement can be further observed as the number of the transformer layer increases from 1 to 12. 
However, performance begins to drop when the number of layers exceeds 12. Likewise, with a fixed number of layer, 
the performance variation on the widths (or heads) of the model follows the same scheme. Based on this result, we set layers to 12, and heads to 8 in all experiments in our main paper.

\begin{table}[!htp]
    \centering
    \setlength\tabcolsep{10pt}
    \resizebox{.6\textwidth}{!}{%
        \begin{tabular}{ccc|cccc|c}
            \toprule
            layers & heads & width & $5^0$ & $5^1$ & $5^2$ & $5^3$ & mIoU \\
            \midrule
            1 &  8 & 512 & 52.2 & 59.2 & 45.0 & 44.6 & 50.2\\
            3 &  8 & 512 & 57.1 & 68.0 & 46.7 & 50.0 & 55.4\\
            6 &  8 & 512 & 58.9 & 68.9 & \textbf{51.3} & 51.5 & 57.7\\
            \underline{12} & \underline{8} &  \underline{512} & \textbf{61.9} & 70.0 & 51.2 & \textbf{52.7} & \textbf{59.0}\\
            18 & 8 &  512 & 57.0 & 69.5 & 43.5 & 49.4 & 54.8\\
            24 & 8 &  512 & 51.1 & \textbf{71.7} & 48.7 & 47.2 & 54.7\\
            \midrule
            12    & 1     & 64    & 54.7 &56.3 &41.4 &43.8 &49.1\\
            12    & 4     & 256   & 60.9 &65.9 &43.2 &45.9 &54.0 \\
            \underline{12}    & \underline{8}     & \underline{512}   & \textbf{61.9} & 70.0 & \textbf{51.2} & \textbf{52.7} & \textbf{59.0} \\
            12 & 10 & 640 & 54.1 & \textbf{70.5} & 49.1 & 47.7 & 55.3\\
            12 & 12 & 768 & 55.5 & 65.8 & 47.8 & 46.8 & 54.0\\
            12 & 16 & 1024 & 51.7 & 66.4 & 49.9 & 49.7 & 54.4\\
            \bottomrule
        \end{tabular}}
    \caption{Performance on layers, heads, and width of cross-modality fusion module on PASCAL-5$^i$. Underlined are the best parameters we used in the main paper.}
    \label{table:architecture-ablation}
\end{table}

\clearpage
\subsection{More Qualitative Visualization}

\begin{figure}[!htp]

  \setlength{\subfigcapskip}{-3pt}
  \setlength{\subfigtopskip}{2pt}
  \vspace{-5pt}
  \setcounter{subfigure}{0}
  \subfigure[image]{
      \centering
      \includegraphics[width=.19\linewidth]{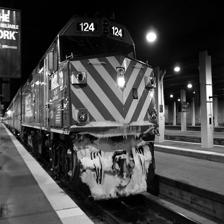}
  }
  \hspace{-7pt}
  \subfigure[image]{
      \centering
      \includegraphics[width=.19\linewidth]{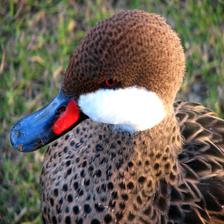}
  }
  \hspace{-7pt}
  \subfigure[image]{
      \centering
      \includegraphics[width=.19\linewidth]{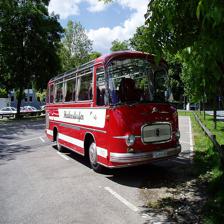}
  }
  \hspace{-7pt}
  \subfigure[image]{
      \centering
      \includegraphics[width=.19\linewidth]{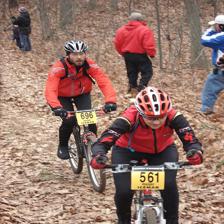}
      \label{fig:pascal_vis_bicycle}
  }
  \hspace{-7pt}
  \subfigure[image]{
      \centering
      \includegraphics[width=.19\linewidth]{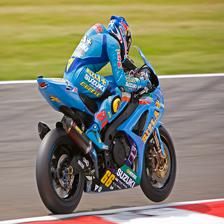}
      \label{fig:pascal_vis_motorbike}
  }\\
  
  \vspace{-5pt}
  \setcounter{subfigure}{0}
  \subfigure[train]{
      \centering
      \includegraphics[width=.19\linewidth]{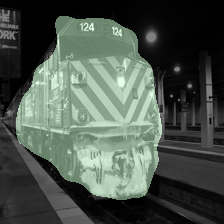}
  }
  \hspace{-7pt}
  \subfigure[bird]{
      \centering
      \includegraphics[width=.19\linewidth]{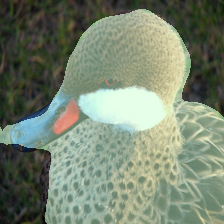}
  }
  \hspace{-7pt}
  \subfigure[bus]{
      \centering
      \includegraphics[width=.19\linewidth]{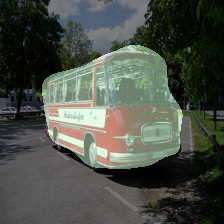}
  }
  \hspace{-7pt}
  \subfigure[bicycle]{
      \centering
      \includegraphics[width=.19\linewidth]{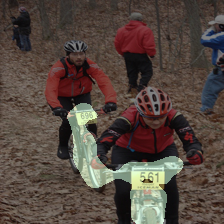}
  }
  \hspace{-7pt}
  \subfigure[motorbike]{
      \centering
      \includegraphics[width=.19\linewidth]{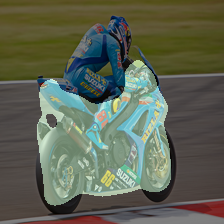}
  }\\
  
  \subfigure[image]{
      \centering
      \includegraphics[width=.19\linewidth]{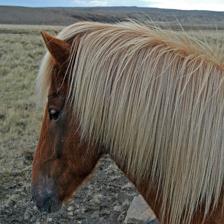}
  }
  \hspace{-7pt}
  \subfigure[image]{
      \centering
      \includegraphics[width=.19\linewidth]{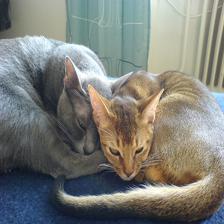}
      \label{fig:pascal_vis_cat}
  }
  \hspace{-7pt}
  \subfigure[image]{
      \centering
      \includegraphics[width=.19\linewidth]{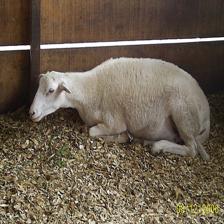}
  }
  \hspace{-7pt}
  \subfigure[image]{
      \centering
      \includegraphics[width=.19\linewidth]{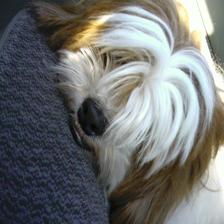}
  }
  \hspace{-7pt}
  \subfigure[image]{
      \centering
      \includegraphics[width=.19\linewidth]{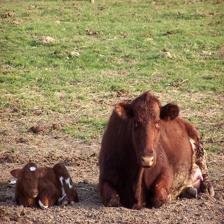}
      \label{fig:pascal_vis_cow}
  }\\
  
  \vspace{-5pt}
  \setcounter{subfigure}{5}
  \subfigure[horse]{
      \centering
      \includegraphics[width=.19\linewidth]{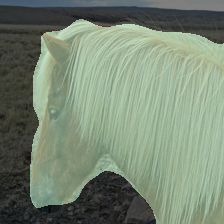}
  }
  \hspace{-7pt}
  \subfigure[cat]{
      \centering
      \includegraphics[width=.19\linewidth]{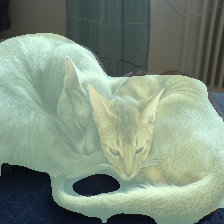}
  }
  \hspace{-7pt}
  \subfigure[sheep]{
      \centering
      \includegraphics[width=.19\linewidth]{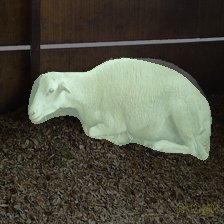}
  }
  \hspace{-7pt}
  \subfigure[dog]{
      \centering
      \includegraphics[width=.19\linewidth]{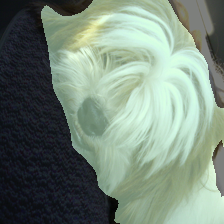}
  }
  \hspace{-7pt}
  \subfigure[cow]{
      \centering
      \includegraphics[width=.19\linewidth]{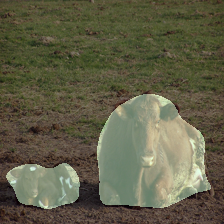}
  }\\
  
  \subfigure[image]{
      \centering
      \includegraphics[width=.19\linewidth]{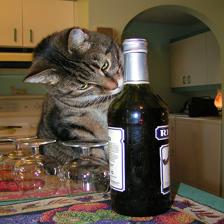}
      \label{fig:pascal_vis_bottle}
  }
  \hspace{-7pt}
  \subfigure[image]{
      \centering
      \includegraphics[width=.19\linewidth]{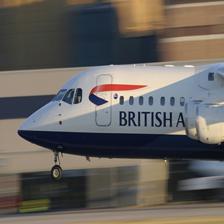}
  }
  \hspace{-7pt}
  \subfigure[image]{
      \centering
      \includegraphics[width=.19\linewidth]{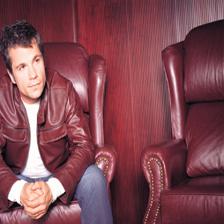}
      \label{fig:pascal_vis_chair}
  }
  \hspace{-7pt}
  \subfigure[image]{
      \centering
      \includegraphics[width=.19\linewidth]{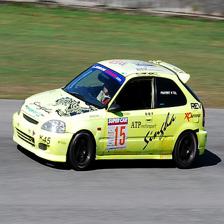}
  }
  \hspace{-7pt}
  \subfigure[image]{
      \centering
      \includegraphics[width=.19\linewidth]{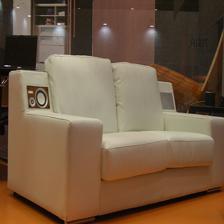}
  }\\
  
  \vspace{-5pt}
  \setcounter{subfigure}{10}
  \subfigure[bottle]{
      \centering
      \includegraphics[width=.19\linewidth]{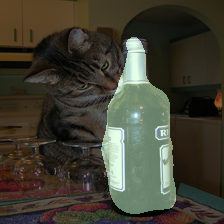}
  }
  \hspace{-7pt}
  \subfigure[aeroplane]{
      \centering
      \includegraphics[width=.19\linewidth]{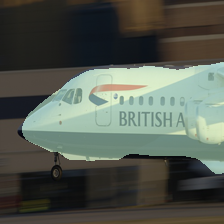}
  }
  \hspace{-7pt}
  \subfigure[chair]{
      \centering
      \includegraphics[width=.19\linewidth]{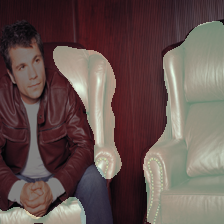}
  }
  \hspace{-7pt}
  \subfigure[car]{
      \centering
      \includegraphics[width=.19\linewidth]{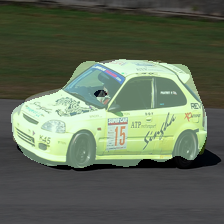}
  }
  \hspace{-7pt}
  \subfigure[sofa]{
      \centering
      \includegraphics[width=.19\linewidth]{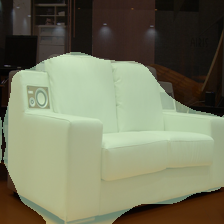}
  }\\
  
  \vspace{-5pt}
  \caption{Qualitative results on PASCAL5$^i$. All categories are novel (unseen) in their corresponding folds, and we show 15 different categories. For each two rows, top: images; bottom: segmentation masks for unseen categories.
  }
  \label{fig:visualize-pascal}
  \vspace{-20pt}
\end{figure}

In this section, we present additional qualitative results on PASCAL-5$^i$ and COCO-20$^i$ using our model with ViT-L backbone. 
Specifically, \figref{fig:visualize-pascal} shows the results on PASCAL-5$^i$. All categories are novel (unseen) in their corresponding fold. Taking into account the variety of images, we choose 15 different categories to display.
\figref{fig:pascal_vis_bicycle} bicycle, 
\figref{fig:pascal_vis_motorbike} motorbike,
\figref{fig:pascal_vis_chair} chair, 
and \figref{fig:pascal_vis_bottle} bottle show \textbf{Fusioner}'s ability to distinguish object of target semantic from other objects~(distractor),
\eg, the person in \figref{fig:pascal_vis_motorbike}, and the cat in \figref{fig:pascal_vis_cat}.
Besides, in \figref{fig:pascal_vis_bicycle} bicycle, \figref{fig:pascal_vis_cow} cow, \figref{fig:pascal_vis_chair} chair, and \figref{fig:pascal_vis_cat} cat, the model segments precisely even when the target instance contains more than one. 

\begin{figure}[!htp]
\setlength{\subfigcapskip}{-3pt}
\setlength{\subfigtopskip}{2pt}
  \vspace{-5pt}
  \setcounter{subfigure}{0}
  \subfigure[image]{
      \centering
      \includegraphics[width=.19\linewidth]{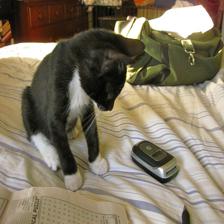}
  }
  \hspace{-7pt}
  \subfigure[handbag*]{
      \centering
      \includegraphics[width=.19\linewidth]{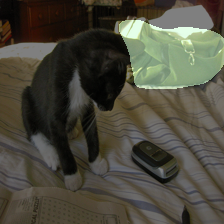}
  }
  \hspace{-7pt}
  \subfigure[bed]{
      \centering
      \includegraphics[width=.19\linewidth]{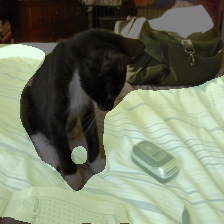}
  }
  \hspace{-7pt}
  \subfigure[cat]{
      \centering
      \includegraphics[width=.19\linewidth]{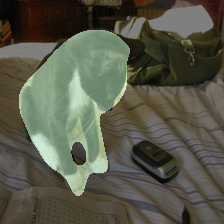}
  }
  \hspace{-7pt}
  \subfigure[cellphone]{
      \centering
      \includegraphics[width=.19\linewidth]{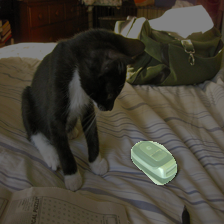}
  }

  \vspace{-5pt}
  \setcounter{subfigure}{0}{
      \subfigure[image]{
          \centering
          \includegraphics[width=.19\linewidth]{images/visualize/000000222681/img.jpg}
      }
      \hspace{-7pt}
      \subfigure[tvmonitor*]{
          \centering
          \includegraphics[width=.19\linewidth]{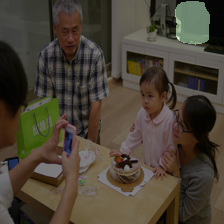}
          \label{fig:coco_vis_tv}
      }
      \hspace{-7pt}
      \subfigure[cake]{
          \centering
          \includegraphics[width=.19\linewidth]{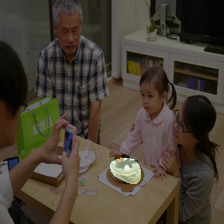}
      }
      \hspace{-7pt}
      \subfigure[diningtable]{
          \centering
          \includegraphics[width=.19\linewidth]{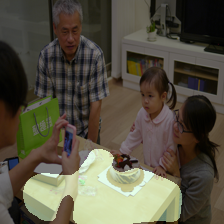}
      }
      \hspace{-7pt}
      \subfigure[person]{
          \centering
          \includegraphics[width=.19\linewidth]{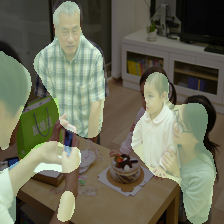}
      }}

  \vspace{-5pt}
  \setcounter{subfigure}{0}
  \subfigure[image]{
      \centering
      \includegraphics[width=.19\linewidth]{images/visualize/000000177213/img.jpg}
  }
  \hspace{-7pt}
  \subfigure[pizza*]{
      \centering
      \includegraphics[width=.19\linewidth]{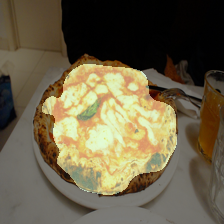}
  }
  \hspace{-7pt}
  \subfigure[cup*]{
      \centering
      \includegraphics[width=.19\linewidth]{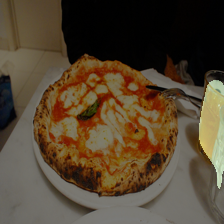}
  }
  \hspace{-7pt}
  \subfigure[diningtable]{
      \centering
      \includegraphics[width=.19\linewidth]{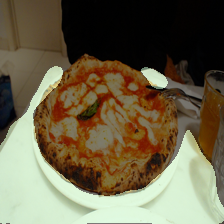}
  }
  \hspace{-7pt}
  \subfigure[person]{
      \centering
      \includegraphics[width=.19\linewidth]{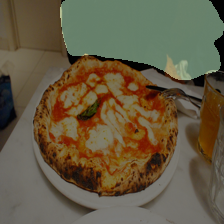}
  }

  \vspace{-5pt} 
  \setcounter{subfigure}{0}
  \subfigure[image]{
      \centering
      \includegraphics[width=.19\linewidth]{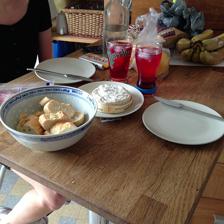}
  }
  \hspace{-7pt}
  \subfigure[banana*]{
      \centering
      \includegraphics[width=.19\linewidth]{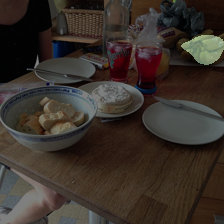}
      \label{fig:coco_vis_banana}
  }
  \hspace{-7pt}
  \subfigure[bowl]{
      \centering
      \includegraphics[width=.19\linewidth]{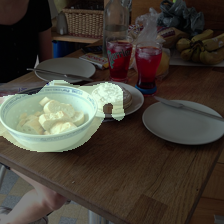}
  }
  \hspace{-7pt}
  \subfigure[cup]{
      \centering
      \includegraphics[width=.19\linewidth]{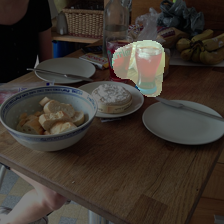}
  }
  \hspace{-7pt}
  \subfigure[diningtable]{
      \centering
      \includegraphics[width=.19\linewidth]{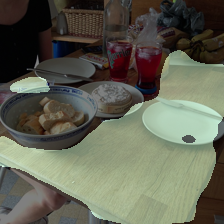}
  }
  
  \vspace{-5pt}
  \setcounter{subfigure}{0}
  \subfigure[image]{
      \centering
      \includegraphics[width=.19\linewidth]{images/visualize/000000115846/img.jpg}
  }
  \hspace{-7pt}
  \subfigure[remote*]{
      \centering
      \includegraphics[width=.19\linewidth]{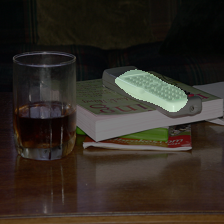}
      \label{fig:coco_vis_remote}
  }
  \hspace{-7pt}
  \subfigure[cup*]{
      \centering
      \includegraphics[width=.19\linewidth]{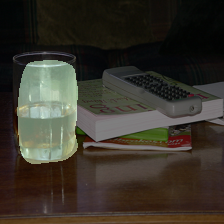}
  }
  \hspace{-7pt}
  \subfigure[diningtable]{
      \centering
      \includegraphics[width=.19\linewidth]{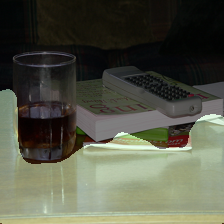}
  }
  \hspace{-7pt}
  \subfigure[sofa]{
      \centering
      \includegraphics[width=.19\linewidth]{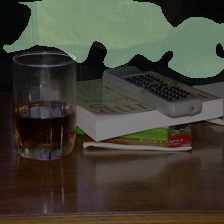}
  }

  \vspace{-5pt}
  \setcounter{subfigure}{0}{
      \subfigure[image]{
          \centering
          \includegraphics[width=.19\linewidth]{images/visualize/000000097696/img.jpg}
      }
      \hspace{-7pt}
      \subfigure[mouse*]{
          \centering
          \includegraphics[width=.19\linewidth]{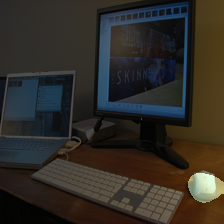}
          \label{fig:coco_vis_mouse}
      }
      \hspace{-7pt}
      \subfigure[keyboard]{
          \centering
          \includegraphics[width=.19\linewidth]{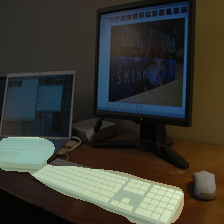}
      }
      \hspace{-7pt}
      \subfigure[laptop]{
          \centering
          \includegraphics[width=.19\linewidth]{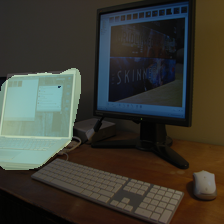}
      }
      \hspace{-7pt}
      \subfigure[tvmonitor]{
          \centering
          \includegraphics[width=.19\linewidth]{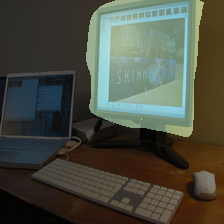}
      }}
  \vspace{-5pt}
  \caption{Qualitative results on COCO-20$^i$. 
  (a) input images, (b)-(e) segmentation masks for different categories. 
  Unseen categories are marked as *.
  }
  \label{fig:visualize-coco}
  \vspace{-20pt}
\end{figure}

The visualization of COCO-20$^i$ is shown in \figref{fig:visualize-coco}, 
with both seen and unseen categories are displayed. 
Facing a more noisy and complex scene, 
\textbf{Fusioner} is still able to recognize the desired (unseen) categories that are small and cornered, for example, tvmonitor, banana, remote and mouse in column of \figref{fig:coco_vis_tv}.

\end{document}